\documentclass{article} 
\usepackage{iclr2027_conference,times}
\iclrfinalcopy

\usepackage{amsmath,amsfonts,bm}

\def\eqref#1{equation~\ref{#1}}

\def\1{\bm{1}}

\DeclareMathAlphabet{\mathsfit}{\encodingdefault}{\sfdefault}{m}{sl}
\SetMathAlphabet{\mathsfit}{bold}{\encodingdefault}{\sfdefault}{bx}{n}

\usepackage{hyperref}
\usepackage{url}
\usepackage{graphicx}
\usepackage{booktabs}
\usepackage{amsmath,amssymb}
\graphicspath{{figures/}}

\usepackage{enumitem}
\usepackage{algorithm}
\usepackage{algpseudocode}
\setlist{topsep=2pt,itemsep=1pt,parsep=0pt,partopsep=0pt,leftmargin=1.4em}
\title{Structure, Dimensionality, and Distribution Shape\\[3pt]
{\large What Makes Pretraining Data Transferable, and Why the Input Pipeline Decides Which}}

\author{{Sangyoon Bae} \\
Interdisciplinary Program in Artificial Intelligence \\
Seoul National University\\
Seoul, 08826, South Korea \\
\texttt{stellasybae@snu.ac.kr} \\
\And
Jiook Cha \\
Department of Psychology \\
Seoul National University \\
Seoul, 08826, South Korea  \\
\texttt{connectome@snu.ac.kr}}

\begin{document}

\maketitle

\begin{abstract}
What makes a pretraining corpus transferable? We describe a corpus by three properties: its
\emph{structure}, its \emph{dimensionality}, and the \emph{distribution shape} of each unit's values. We
propose the \textbf{Transfer-Invariance Principle}: the same property can be irrelevant to transfer under
one input transform and decisive under another, so the model's input transform, not the data modality,
decides which properties can matter. To test it we build \textbf{TransferForge}, a generator that places a
synthetic population at any prescribed structure, dimensionality, and distribution shape. KernelSynth, the
synthetic prior of Chronos, cannot, because it draws independent series. When the model z-scores each input window (PatchTST on
UCR), the distribution shape is inert and the structural setting governs. Synthetic pretraining then
matches a real corpus of the same size and exceeds it once scaled. When the model encodes calcium $\Delta F/F$
unnormalized (POYO+), the shape becomes a governing constraint, and a shape-matched synthetic corpus
replaces real pretraining with \emph{zero real neural activity}. Flipping only the input normalization, with
the model and data fixed, changes the effect of a shape manipulation $\mathbf{5.3\times}$, so the pipeline,
not the modality, sets the switch. The principle yields a recipe that chooses what to synthesize before any
training. Pretraining data then becomes something to manufacture for the model, most valuably where real
data is scarce, expensive, or private.
\end{abstract}

\section{Introduction}
\label{sec:intro}

Time-series foundation models are known to pretrain well entirely on synthetic data \citep{xie2026cauker}.
Pretraining data can therefore be designed to do its job: to teach the model structure that transfers when it
is finetuned on real data. Which property of a corpus achieves this, and whether that holds across models,
is not known.
A partial answer exists for one modality: in calcium imaging, predictable-first pretraining improves
neural transfer \citep{bae2026decoding}. Extending that analysis across modalities
(Appendix~\ref{app:background}), we distill two structural properties of a corpus, \emph{structure}
(how reconstructable a unit is from the rest) and \emph{dimensionality} (the number of independent modes),
which describe how the units relate. A third is the per-unit \emph{distribution shape} (the shape of each
unit's own values, the marginal).

These properties alone do not predict transfer. The same Gaussian synthetic generator gives opposite results
on two targets. Pretraining PatchTST for UCR classification, its corpus matches a real forecasting corpus of
the same size and, scaled up, exceeds it by $4.2\%$. Pretraining POYO+ for calcium decoding, mixed with a small real pool, it transfers $27\%$ worse than no
pretraining, while a calcium-shaped corpus in the same mixture transfers $14\%$ better. Either the target modality or the model decides which
property matters.

\begin{figure}[!ht]
  \centering
  \includegraphics[width=0.75\linewidth]{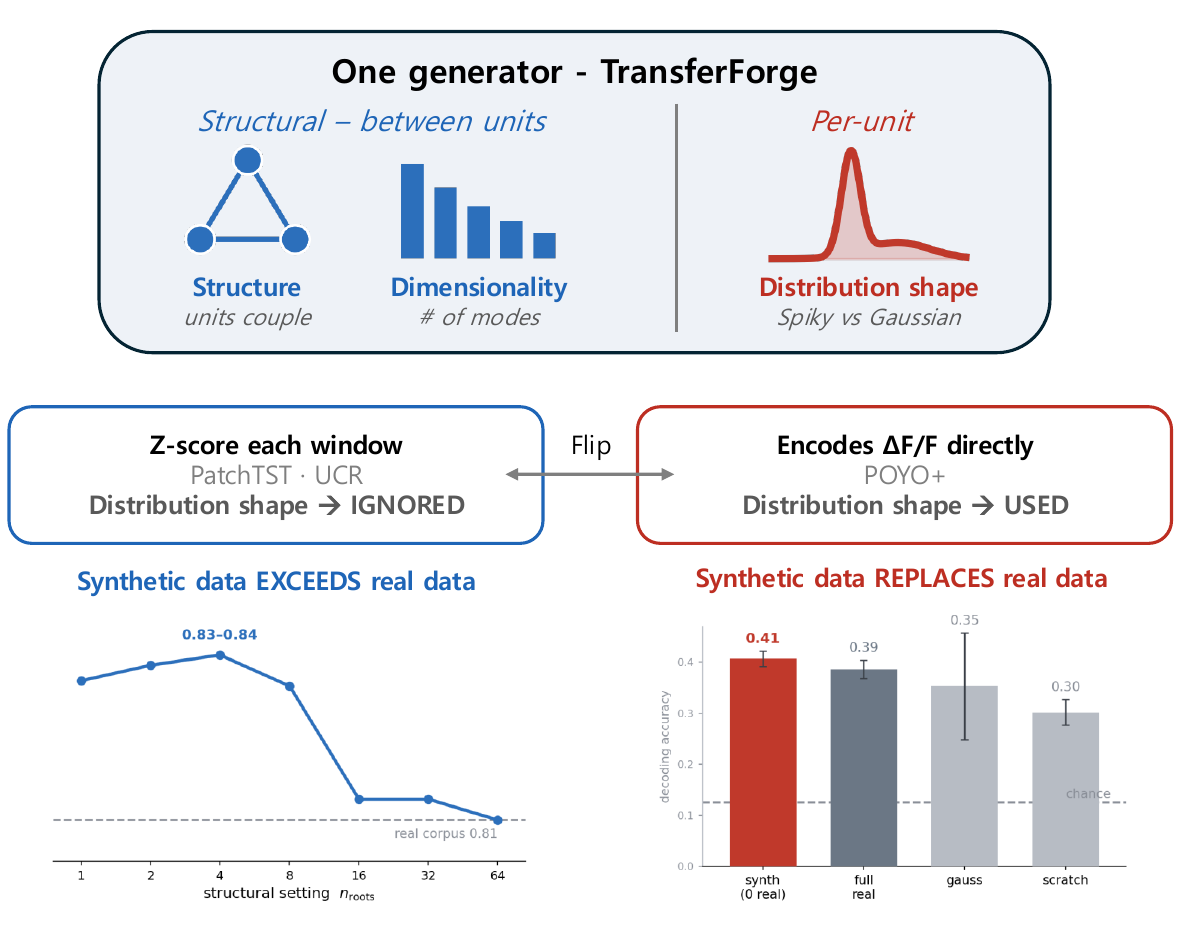}
  \caption{\textbf{The Transfer-Invariance Principle.} A pretraining corpus is a point on three axes (top):
  two \emph{structural} (structure, dimensionality) and the per-unit \emph{distribution shape}. The model's
  input pipeline sets which axes can govern transfer (middle). Under per-window normalization, synthetic data
  exceeds a real corpus once scaled (bottom left, PatchTST on UCR). Under unnormalized input, shape-matched
  synthetic data replaces real pretraining (bottom right, POYO+ on calcium imaging).}
  \label{fig:overview}
  \vspace{-8pt}
\end{figure}

We show that the model decides, through its input pipeline. PatchTST z-scores each input window; POYO+
encodes $\Delta F/F$ unnormalized. We call this the \textbf{Transfer-Invariance Principle}
(Figure~\ref{fig:overview}): the same corpus property can be irrelevant to transfer under one input transform and decisive under
another, so the model's input transform, not the data modality, decides which properties can matter. Flipping only the input normalization, on one model and one
dataset, reproduces the switch.

To test the principle we build \textbf{TransferForge}, a generator that places a population at any
prescribed structure and dimensionality. A few independent latent sources set the dimensionality ceiling,
and a controllable fraction of derived, reconstructable nodes sets embeddedness. Together they span the
(effective rank, coupling) plane of real neural populations (Section~\ref{sec:method}).

Our contributions are:
\begin{itemize}
  \item \textbf{Which data property transfers depends on the model's input pipeline.} Flipping only the
    input normalization changes the measured effect of the distribution-shape manipulation $5.3\times$.
  \item \textbf{Structure, dimensionality, and distribution shape characterize transferable pretraining
    data.} Under per-window normalization, the structural axes govern, with an interior optimum.
    Under unnormalized input, the distribution shape becomes a governing constraint.
  \item \textbf{A generator that sets each property on its own.} TransferForge places a synthetic
    population anywhere from coupling $0.06$ to $0.67$ and effective rank $1.8$ to $41$, and matches a target
    distribution shape separately. KernelSynth, which draws independent series, occupies one point on these
    axes and cannot make a coupled population.
  \item \textbf{Synthesizing along the governing property helps real-data finetuning.} A shape-matched
    corpus replaces full-real calcium pretraining with zero real neural activity ($0.407$ vs.\ $0.386$), and
    a structure-set corpus matches a real corpus of equal size on UCR and exceeds it once scaled up.
\end{itemize}

\section{Related work}
\label{sec:related}

\paragraph{Synthetic pretraining data.} Time-series and tabular foundation models are increasingly pretrained on hand-designed priors:
KernelSynth in Chronos \citep{ansari2024chronos}, ForecastPFN \citep{dooley2023forecastpfn}, TabPFN
\citep{hollmann2023tabpfn}, and domain randomization \citep{tobin2017domain}. CauKer
\citep{xie2026cauker}, a Gaussian-process-plus-structural-causal-model (GP$+$SCM) generator, is closest. It
shows that classification time-series foundation models can be pretrained \emph{purely} on synthetic data.
We start from this result. CauKer establishes that a synthetic corpus can work, and we ask which property of
that corpus makes it transfer and what decides that property. To answer this, TransferForge extends the
GP$+$SCM construction of CauKer so that structure, dimensionality, and distribution shape can each be set on
their own (Section~\ref{sec:method}).

\paragraph{Choosing training data.} Which training data helps a model has also been studied from several angles. Curriculum learning orders
examples from easy to hard \citep{bengio2009curriculum}. Data pruning and online batch selection keep the
examples most worth training on, to improve accuracy or speed on the same task
\citep{sorscher2022beyond,mindermann2022prioritized}. Both act on data that already exist. Neither
identifies which corpus property makes pretraining transfer. We identify that property and synthesize data
that has it. Remaining connections (population coupling, intrinsic dimension, scaling) are deferred to
Appendix~\ref{app:related}.

\section{Methods}
\label{sec:method}

\subsection{Three corpus properties}
\label{sec:axes}

\paragraph{Structure and dimensionality.} We describe a population $X\in\mathbb{R}^{T\times N}$ by two
structural axes, its \emph{structure} (embeddedness) and its \emph{dimensionality} (effective rank):
\begin{equation}
  \underbrace{\mathrm{coup}(X)=\operatorname*{mean}_{i\neq j}\,|\mathrm{corr}(x_i,x_j)|}_{\text{structure (embeddedness)}}
  \qquad\qquad
  \underbrace{\mathrm{er}(X)=\frac{\big(\sum_k\lambda_k\big)^2}{\sum_k\lambda_k^2}}_{\text{dimensionality (effective rank)}}
\end{equation}
Structure is the mean absolute off-diagonal correlation. It summarizes at the population level how well
each unit is reconstructed from the rest, which Appendix~\ref{app:background} also reports per unit as a
cross-validated $R^2$. Dimensionality is the participation ratio of the correlation eigenspectrum
$\{\lambda_k\}$ and counts the independent modes. Within a real recording, structure and dimensionality are
strongly anticorrelated: across random neuron pools from one \textit{C.\ elegans} whole-brain session,
$\rho\approx-0.85$. Most of the plane therefore cannot be reached by collecting real data, and TransferForge
is built to span it. Both axes come from a prior cross-modality analysis summarized in
Appendix~\ref{app:background}.

\paragraph{Distribution shape.} The third axis is the per-unit \emph{distribution shape}, the shape of each
unit's own values. We summarize it by skewness, excess kurtosis, and the peak-to-baseline signal-to-noise
ratio standard for calcium traces,
\begin{equation}
  \mathrm{SNR}(x) \;=\; \frac{\mathrm{P}_{99}(x) - \operatorname{median}(x)}
       {\operatorname{std}\!\big(x < \operatorname{median}(x)\big)},
\end{equation}
the transient amplitude over the baseline noise. The $99$th percentile estimates the peak without being
driven by single-sample spikes, the median estimates the baseline, and the standard deviation of the
below-median samples estimates the baseline noise. Structure and dimensionality describe how units
\emph{relate} to each other. The distribution shape describes each unit's own \emph{values}. The model's
input pipeline decides whether the shape must be matched (Section~\ref{sec:marginal}).

\subsection{Real modalities on the three axes}
\label{sec:modalities}

Real modalities occupy very different regions of the three
axes (Table~\ref{tab:modalities}; data sources in Appendix~\ref{app:modalities}). Human macro-scale data
(fMRI, EEG) has low rank and high coupling. Cellular and sensor data spread across the plane. Only calcium,
and to a lesser degree spikes, is heavy-tailed. Every other time-ordered modality is near-Gaussian. A
generator must therefore cover far more than any single modality occupies. Section~\ref{sec:recipe} uses
the table to predict the governing axis for modalities we did not test.

\begin{table}[!ht]
  \centering
  \caption{\textbf{The three axes measured across modalities.} Tail and SNR are computed per unit and
  summarized by the median. They are defined only for time-ordered data (vision and language: ``--'').}
  \label{tab:modalities}
  \begin{tabular}{lcccc}
    \toprule
    & & & \multicolumn{2}{c}{Distribution shape} \\
    \cmidrule(lr){4-5}
    Modality & \shortstack{Structure\\(embeddedness)} & \shortstack{Dimensionality\\(effective rank)} & \shortstack{Tail\\(excess kurtosis)} & SNR \\
    \midrule
    fMRI (task)       & 0.54  & 2.6  & $-0.4$ & 3.4 \\
    EEG               & 0.47  & 3.2  & $0.1$  & 3.7 \\
    fMRI (resting)    & 0.40  & 4.3  & $-0.1$ & 3.9 \\
    weather           & 0.32  & 31   & $0.4$  & 3.5 \\
    \textit{C.\ elegans} & 0.28 & 7.0 & $-0.1$ & 6.5 \\
    vision            & 0.21  & 40   & --     & -- \\
    spikes            & 0.08  & 54   & $3.7$  & 12.9 \\
    calcium           & 0.04  & 50   & $12.0$ & 5.6 \\
    language          & 0.019 & 1063 & --     & -- \\
    \bottomrule
  \end{tabular}
\end{table}

\subsection{The TransferForge generator}
\label{sec:generator}

\paragraph{Construction.} TransferForge adapts the structural-causal-model construction over
Gaussian-process sources of CauKer \citep{xie2026cauker} and makes it \emph{axis-controlled}. We draw
$n_{\text{roots}}$ independent latent source signals, which set the ceiling on dimensionality. We then form
$N - n_{\text{roots}}$ \emph{derived} nodes. Each derived node applies an activation to a random linear mix
of earlier nodes and adds a private-noise term:
\begin{equation}
  x_j \;=\; \phi\!\Big(\sum_{p \in \mathrm{pa}(j)} w_{jp}\, x_p + b_j\Big)
  \;+\; \underbrace{\mathrm{priv}\cdot\sigma(x_j)\,\eta_j}_{\text{private noise}},
\end{equation}
where $\mathrm{pa}(j)$ is a random subset of earlier nodes and $\eta_j$ is white noise. A derived node
is reconstructable from its parents by construction, so it is \emph{embedded}. The fraction of derived
nodes sets embeddedness, and the private-noise weight $\mathrm{priv}$ sets coupling \emph{strength}
independently of that fraction. With $n_{\text{roots}}=1$ and $\mathrm{priv}=0$ the population is close to rank one.
Raising $n_{\text{roots}}$ or $\mathrm{priv}$ makes the population richer (implementation details in
Appendix~\ref{app:impl}).

\paragraph{The structural setting spans the real plane.} We call the pair $(n_{\text{roots}},
\mathrm{priv})$ the \emph{structural setting}. Sweeping it covers the (effective rank, coupling) plane rather
than a single diagonal. Across the grid, effective rank and coupling correlate at $\rho\approx -0.87$, close
to the real within-recording value. The generator parameters are controls, not axis definitions. Every
analysis uses the structure and effective rank measured on the built corpus. An existing synthetic prior
cannot be steered this way. KernelSynth draws each series independently, so a
population of $128$ KernelSynth series occupies one point (coupling $0.075$, effective rank $35$). At the same
size, TransferForge spans an eleven-fold range in coupling and a twenty-three-fold range in effective rank,
and sets the distribution shape separately.

\subsection{Templating into a real model's data format}
\label{sec:templating}

To pretrain an existing model on synthetic data
without changing its loader, we copy a real session file and overwrite \emph{only} its neural traces. The
replacement is a TransferForge population of the same shape $(T,N)$, amplitude-matched per unit to the real
column it replaces. The file keeps every field the loader needs: domain, stimulus intervals, unit ids,
sampling rate, and masks. No real neural \emph{activity} remains. Only the file format, the stimulus
metadata, and each unit's mean and standard deviation come from the real session. A split file marks the
synthetic sessions as the pretraining pool and the untouched real sessions as the finetuning pool.

\section{Results}
\label{sec:results}

\subsection{Datasets and setup}
\label{sec:setup}

\textbf{Two pipelines.} We evaluate transfer under the two input pipelines the principle distinguishes. The
\emph{normalizing} pipeline is a PatchTST encoder that z-scores each input window. The \emph{unnormalized} pipeline
is POYO+ \citep{azabou2025multi}, a calcium population model that consumes $\Delta F/F$ without per-window
normalization. Chronos, a second backbone with a different tokenizer, tests the
principle on another architecture (Section~\ref{sec:gate}).

\textbf{Downstream tasks.} For the normalizing pipeline, we pretrain on a source corpus and train a frozen
logistic-regression probe on each of $22$ datasets of the UCR archive \citep{dau2019ucr}. We report accuracy
averaged over the datasets. For the unnormalized pipeline, we pretrain and then fully finetune an 8-way
drifting-gratings orientation decoder on a held-out pool of $139$ real calcium sessions. We report held-out
test accuracy (chance $0.125$).

\textbf{Pretraining corpora.} Every synthetic corpus is a TransferForge population placed at a prescribed
structure, dimensionality, and distribution shape (Section~\ref{sec:method}). We compare each against the
real corpus a practitioner would use in that pipeline: a real forecasting corpus for UCR, and full real
calcium pretraining for the decoder. Random init, and chance for the decoder, bound every comparison from
below. Corpora compared directly are budget- or step-matched, so the corpus is the only thing that varies. The
scaling analysis varies corpus size on purpose.

\textbf{Why these datasets.} UCR and calcium imaging lie at opposite ends of the input-pipeline spectrum.
UCR is the standard univariate classification archive, and its $22$ datasets span many domains. The real
forecasting corpus, FRED-MD macroeconomic series \citep{mccracken2016fredmd,godahewa2021monash} and daily
exchange rates \citep{lai2018modeling}, is disjoint from UCR, so pretraining cannot leak the evaluation sets.
Calcium $\Delta F/F$ is sparse and heavy-tailed, and POYO+ takes it unnormalized. Real recordings are costly,
so replacing them matters most. We decode drifting-grating orientation from the Allen Brain Observatory
\citep{devries2020large}.

\subsection{Structure governs under per-window normalization (PatchTST on UCR)}
\label{sec:ucr}

We first ask \emph{which} property of the corpus governs transfer for a generic time-series model. We
pretrain a PatchTST encoder and probe transfer on the UCR archive (setup in Section~\ref{sec:setup}, budget in
Appendix~\ref{app:expdetails}).

The structural setting governs transfer, and it has an interior optimum. Here $n_{\text{roots}}$ is the number
of independent latent sources, which largely sets the effective rank. As $n_{\text{roots}}$ increases, transfer
rises, peaks at $n_{\text{roots}}=4$, and falls again (Figure~\ref{fig:overview}). The high-dimensional extreme transfers clearly worse than the middle, and the rank-one
extreme slightly worse. At a fixed corpus size, the interior optimum ($n_{\text{roots}}=4$) improves UCR accuracy by $4\%$ over both the
high-dimensional extreme and random init. KernelSynth draws each series independently and has no parameter that sets
structure, so it cannot make this choice.

Pretrained on the synthetic corpus, PatchTST achieves $4.2\%$ higher UCR accuracy than with the real
forecasting corpus (Table~\ref{tab:ucr}). The real corpus itself transfers no better than training from
scratch. The gain comes from quantity. Cut to the $1776$ windows of the real corpus and trained for the same
steps, the synthetic corpus matches the real one (paired $t$-test over seeds, $p=0.60$). At about ten thousand
windows, our corpus outperforms KernelSynth, the synthetic prior of Chronos, by $1.3\%$ ($8$ of $10$ seeds; paired $t$-test, $p=0.015$). The margin is not uniform across datasets ($13$ of
$22$; Wilcoxon, $p=0.07$).

\begin{table}[!ht]
  \centering
  \caption{\textbf{UCR classification accuracy after pretraining.} PatchTST is pretrained on each corpus,
  then a frozen probe is trained on each UCR dataset; mean $\pm$ SD over 3 seeds (10 for the $\sim$10k synthetic
  rows), averaged over 22 UCR
  datasets. Ours is TransferForge at the interior optimum ($n_{\text{roots}}=4$). Rows with the same
  window count pretrain for the same number of steps; $1776$ is the size of the real corpus.}
  \label{tab:ucr}
  \begin{tabular}{lc}
    \toprule
    Pretraining corpus & Classification accuracy \\
    \midrule
    \multicolumn{2}{l}{\textit{Synthetic}} \\
    \quad \textbf{Ours} ($\sim$10k windows)                   & \textbf{0.842 $\pm$ 0.006} \\
    \quad KernelSynth ($\sim$10k windows) & 0.831 $\pm$ 0.008 \\
    \quad Ours ($1776$ windows)                               & 0.813 $\pm$ 0.014 \\
    \quad KernelSynth ($1776$ windows) & 0.808 $\pm$ 0.003 \\
    \midrule
    \multicolumn{2}{l}{\textit{Real}} \\
    \quad FRED-MD $+$ exchange rate ($1776$ windows)          & 0.808 $\pm$ 0.013 \\
    \midrule
    No pretraining                                            & 0.805 $\pm$ 0.008 \\
    \bottomrule
  \end{tabular}
\end{table}

\begin{figure}[!ht]
  \centering
  \includegraphics[width=0.6\linewidth]{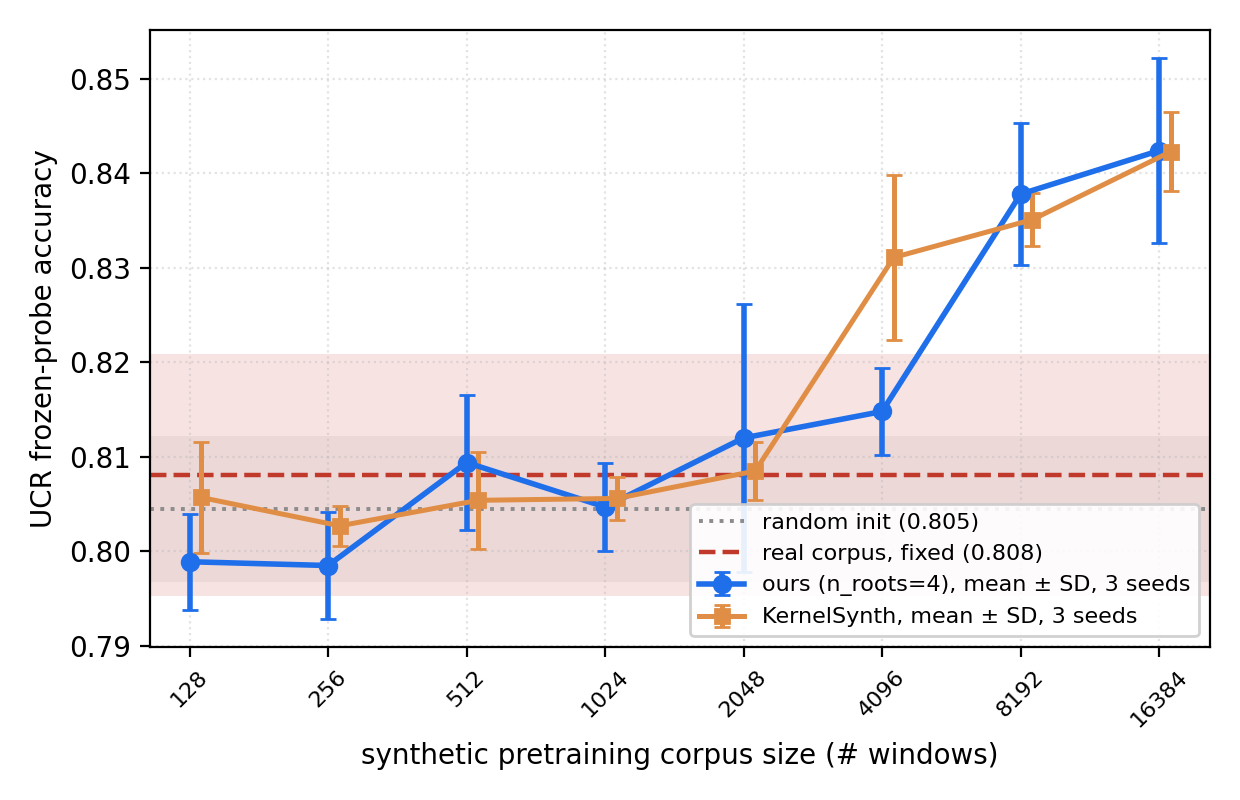}
  \caption{\textbf{UCR classification accuracy vs.\ synthetic corpus size.} PatchTST, frozen probe, at the
  interior optimum ($n_{\text{roots}}=4$); mean $\pm$ SD over 3 seeds, averaged over 22 UCR
  datasets. Blue: our corpus. Orange: KernelSynth at the same sizes and seeds. Dashed: the fixed real corpus
  ($1776$ windows). Dotted: random init.}
  \label{fig:scaling}
\end{figure}

\paragraph{Synthetic transfer scales with corpus size.} A synthetic corpus can grow where a real one is
fixed. At the interior optimum ($n_{\text{roots}}=4$), UCR accuracy stays at the level of the real corpus up to about four thousand
windows and then exceeds it (Figure~\ref{fig:scaling}). At $16384$ windows it is $4.2\%$ higher. KernelSynth
rises the same way, somewhat earlier, and ends at the same accuracy, so the gain from scale comes from synthetic data in
general. Our generator decides what to synthesize, and corpus size separately decides how much the corpus helps.

\subsection{Distribution shape becomes a governing constraint under unnormalized input (POYO+ on calcium imaging)}
\label{sec:neural}

We next test replacement and supplementation on calcium imaging data with POYO+ \citep{azabou2025multi}
(setup in Section~\ref{sec:setup}, budget in Appendix~\ref{app:expdetails}). The rows of Table~\ref{tab:neural} differ only in the
pretraining pool composition. All synthetic corpora use $n_{\text{roots}}=64$. Under the calcium transform the
choice of $n_{\text{roots}}$ barely changes decoding (Section~\ref{sec:effrank}), so we fix it rather than
tune it.

\begin{table}[!ht]
  \centering
  \caption{\textbf{Drifting-gratings decoding accuracy after pretraining.} POYO+, 15-epoch full finetune; held-out
  test accuracy, mean $\pm$ SD. Chance $=0.125$. Rows are over 3 seeds, except the three rows that use
  scarce real data (5 seeds each, including chance-stalled runs, which widens their spread).}
  \label{tab:neural}
  \begin{tabular}{lc}
    \toprule
    Pretraining corpus & Decoding accuracy \\
    \midrule
    \multicolumn{2}{l}{\textit{Synthetic only (0 real traces)}} \\
    \quad \textbf{Calcium-shaped synthetic}         & \textbf{0.407 $\pm$ 0.015} \\
    \quad Gaussian synthetic                        & 0.353 $\pm$ 0.105 \\
    \midrule
    \multicolumn{2}{l}{\textit{Synthetic $+$ scarce real}} \\
    \quad Calcium-shaped synthetic $+$ scarce real  & 0.343 $\pm$ 0.052 \\
    \quad Gaussian synthetic $+$ scarce real        & 0.220 $\pm$ 0.106 \\
    \midrule
    \multicolumn{2}{l}{\textit{Real only}} \\
    \quad Full real                                 & 0.386 $\pm$ 0.018 \\
    \quad Scarce real                               & 0.239 $\pm$ 0.114 \\
    \midrule
    No pretraining                                  & 0.302 $\pm$ 0.025 \\
    \bottomrule
  \end{tabular}
\end{table}

Shape-matched synthetic pretraining replaces real pretraining. With no real neural data, the
calcium-shaped synthetic corpus scores $5\%$ higher than full-real pretraining, a difference that is not
significant ($p=0.20$, Welch). It scores $35\%$ higher than no pretraining ($p=0.007$, Welch;
Table~\ref{tab:neural}). Like full-real pretraining, it rises above chance on every seed, and its spread is
the smallest in the table.

Synthetic data also supplements a small real pool. On their own, $19$ real sessions fall $21\%$ below no
pretraining. Adding $50$ calcium-shaped synthetic sessions improves them by $44\%$. The distribution shape
matters most when synthetic data is mixed with real data. With the same scarce real pool, the calcium-shaped
corpus transfers $56\%$ better than the Gaussian one, which falls below no pretraining
(Table~\ref{tab:neural}). Without real data, the Gaussian corpus also transfers above no pretraining on
average, but only on some seeds. Two of its seeds match the calcium-shaped corpus, and one rises above
chance only late in finetuning. The calcium-shaped corpus succeeds on all three seeds with a seven-fold
smaller spread. The Gaussian corpus is not structureless. It comes from the same generator setting, and
because it skips the calcium transform it keeps more of that structure. At the model's input window its
coupling is $60\%$ higher than that of the calcium-shaped corpus. Its success on two seeds may come from this
retained structure, which we do not isolate here. Every row shares the same finetuning, so these differences come from pretraining
(Appendix~\ref{app:expdetails}).

\paragraph{Why the distribution shape matters here.} POYO+ takes \emph{unnormalized} $\Delta F/F$ values through a
patch of size one and a plain linear value embedding. Real $\Delta F/F$ is sparse and spiky. Over its full
trace, the session the synthetic corpus is calibrated on has median skew $\approx 6.8$ and median excess
kurtosis $\approx 97$ across units. A Gaussian TransferForge population, matched only in mean and standard
deviation, has skew and excess kurtosis near zero. Without normalization these higher moments reach the
model unchanged, and the mismatch enters as a distribution shift.

The transient-to-baseline SNR must also match the real level ($\approx 6.8$ for the calibration session, and
$5.6$ for the calcium median in Table~\ref{tab:modalities}). Our calcium distribution-shape transform maps a
TransferForge latent to a calcium-like signal. It brings skewness, excess kurtosis, and SNR to the level of
the calibration session at once (Figure~\ref{fig:marginal}; construction in Appendix~\ref{app:impl}). The
result is somewhat heavier-tailed than the calibration session and than a typical session (excess kurtosis
$\approx 140$).

\paragraph{Which component of the distribution shape matters.} At the structural setting of all calcium
corpora ($n_{\text{roots}}=64$), we vary how the per-unit distribution is matched: Gaussian, shape only, SNR only, or
both. Matching both gives the highest and most reliable decoding (Figure~\ref{fig:calcium_axes}c). Matching shape
only or SNR only falls below no pretraining, and below the Gaussian corpus. In calcium,
the transform changes shape, SNR, and effective rank together, so Table~\ref{tab:neural} credits the
transform as a whole. On Chronos, whose inputs are dense series, the tail shape carries the effect and SNR
matters only near the noise floor (Section~\ref{sec:gate}).

\subsection{Once the heavy tail is matched, the structural axes do not improve transfer (POYO+ on calcium imaging)}
\label{sec:effrank}

\begin{figure}[!ht]
  \centering
  \includegraphics[width=\linewidth]{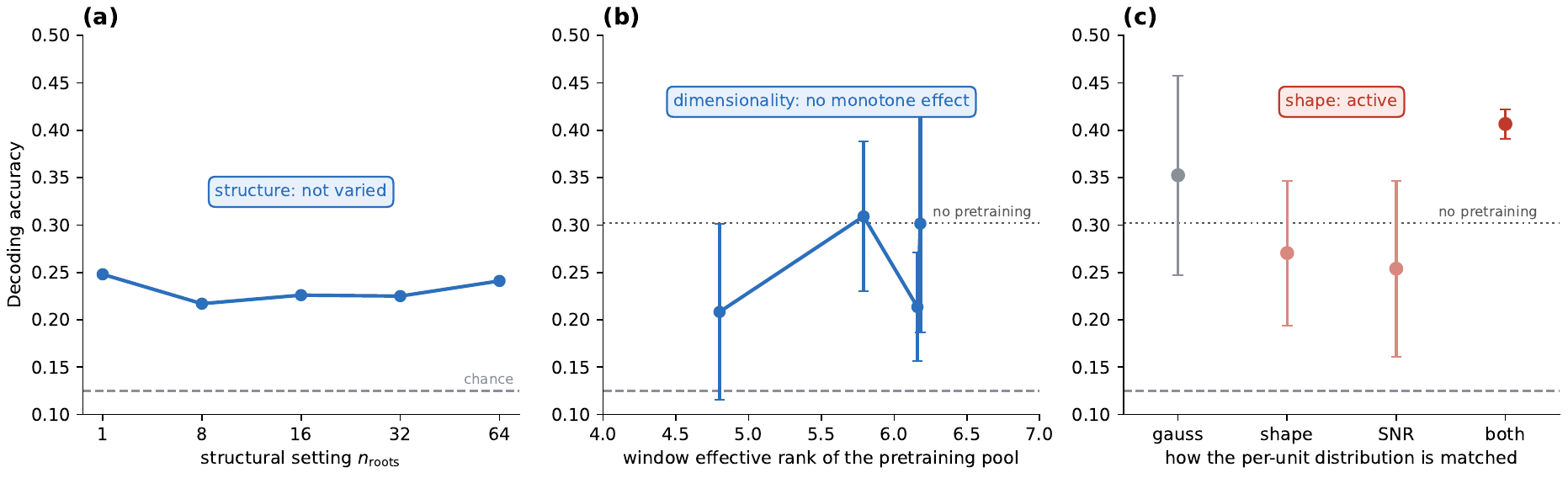}
  \caption{\textbf{Calcium drifting-gratings decoding accuracy (chance $0.125$) as one property is varied.} (a)~structural setting ($n_{\text{roots}}$), (b)~window effective rank of the
  pretraining pool (real sessions span $2.1$--$16.8$, 5th--95th percentile), (c)~per-unit distribution shape match (gauss/shape/SNR/both). Held-out test accuracy. The
  distribution shape is re-matched in (a) and (b). (a)~is single-seed at a 5-epoch budget, so its levels are
  not comparable to the other panels; (b) and (c) show mean $\pm$ SD over 3 seeds after
  10-epoch pretraining; drifting-gratings finetuning ran 15 epochs, except runs that hit the time limit and
  were rerun at 12 (7 of the 12 runs in (b); the seed-43 and seed-44 runs of shape and SNR in (c)). The dotted line
  is the 15-epoch no-pretraining level.}
  \label{fig:calcium_axes}
\end{figure}

\paragraph{Matching the calcium shape erases the structure we set.} We set structure with $n_{\text{roots}}$
from $1$ to $64$ and then applied the calcium transform, and decoding stays flat
(Figure~\ref{fig:calcium_axes}a). The transform explains this. Before it, coupling changes more than
fifty-fold across $n_{\text{roots}}$. After it, coupling changes less than two-fold, within the variation
between real sessions (Figure~\ref{fig:neural_sweep}). The corpora the model sees therefore differ little in
structure, and the flat sweep does not show that structure is irrelevant. The transform cannot be dropped,
because POYO+ encodes values unnormalized and the corpus must carry the calcium shape
(Section~\ref{sec:neural}). Matching the shape and varying structure therefore conflict
(Appendix~\ref{app:expdetails}).

\paragraph{Effective rank, varied directly, shows no systematic effect.} We therefore varied effective rank
directly through the event density $q$ of the transform. At each level we re-matched the distribution shape
and kept coupling fixed across levels (Appendix~\ref{app:expdetails}). Varying $q$ moves effective rank only
within a narrow band, about a tenth of the range of real sessions (Figure~\ref{fig:calcium_axes}b). Within that
band decoding shows no systematic trend and stays within the seed spread of each level. Every level is
shape-matched, yet none rises clearly above no pretraining, and all stay well below the default shape-matched corpus
(Figure~\ref{fig:calcium_axes}c). Matching the distribution shape is therefore not sufficient on its own.

\subsection{The pipeline, not the modality, decides which properties can matter (PatchTST and Chronos on time series)}
\label{sec:gate}\label{sec:marginal}

\begin{figure}[!ht]
  \centering
  \includegraphics[width=\linewidth]{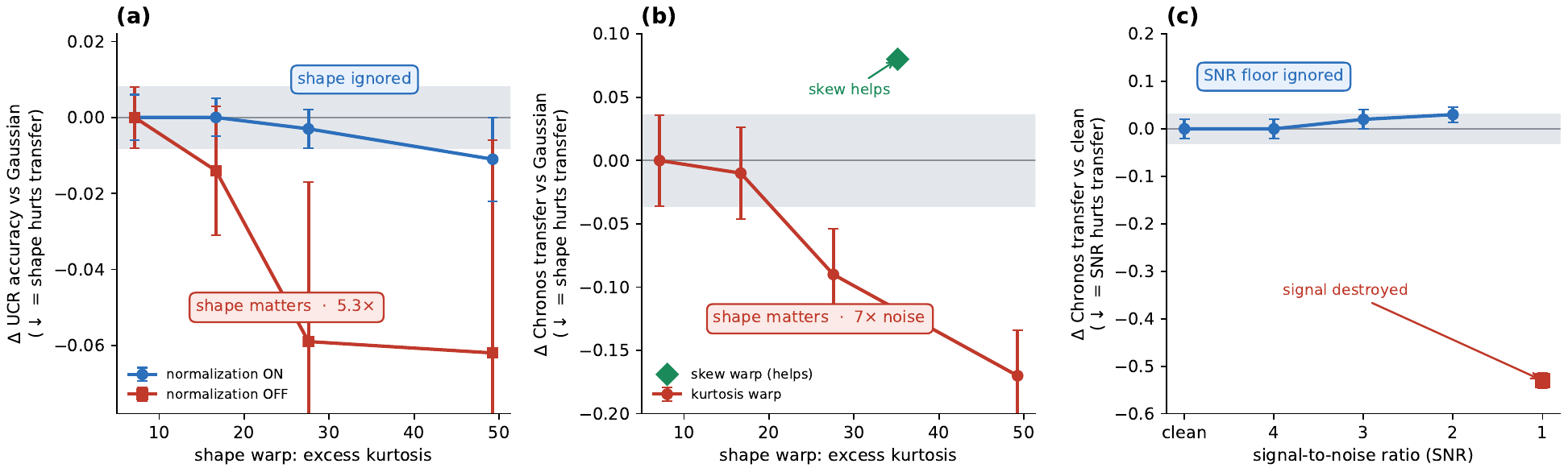}
  \caption{\textbf{Distribution-shape effect with and without per-window normalization.} Each panel plots the transfer change from the Gaussian anchor, oriented so
  \emph{down $=$ shape hurts}. The grey band is within seed noise. (a)~PatchTST, per-window normalization
  on vs.\ replaced by a corpus-wide scale (10 seeds). (b)~Chronos with its native tokenizer, no
  intervention (3 seeds). (c)~Chronos, an SNR sweep with tail shape held Gaussian (3 seeds). Panels use
  different tasks and metrics.}
  \label{fig:gate}
\end{figure}

UCR and calcium differ in modality as well as pipeline, so the contrast between them could still reflect
the data rather than the normalization. We therefore fix the model and the data and flip only the input
normalization.

On PatchTST at the interior optimum ($n_{\text{roots}}=4$), we warp one synthetic corpus with a pointwise monotone
map. The map keeps the temporal order of each series and changes only its distribution shape
(post-normalization excess kurtosis from about $7$ to $49$). With per-window normalization on, the warp has no
effect. UCR accuracy stays within seed noise (Figure~\ref{fig:gate}a), while changing the structural setting
moves it by $4\%$ (Section~\ref{sec:ucr}). We then replace per-window normalization with a single corpus-wide scale, so each
window's scale also reaches the model. The same warp then matters. Accuracy falls monotonically with
kurtosis, and the spread grows $5.3$-fold. The model, the data, and the warp stay the same, and only the
normalization changes. Without normalization the warp also changes each window's scale, and this
experiment does not separate shape from scale.

The flip also holds on real data. On real forecasting series (FRED-MD and exchange rates), the warp is again
inert with normalization on and active without it, with a $2.9$-fold larger spread. On a KernelSynth base,
normalization again makes the warp inert. Without normalization the spread
doubles, less than on the other two bases (Appendix~\ref{app:flip}).

This is not a tautology, because z-scoring keeps the higher moments. A second model with a different
tokenizer, Chronos, responds to the warp without any intervention. Its forecasting error rises monotonically
with tail heaviness, a spread seven times the seed noise (Figure~\ref{fig:gate}b). Chronos also separates
the two components of the distribution shape. Lowering SNR from clean down to $2$ leaves forecasting flat,
and it degrades only at SNR $1$, where the signal variance equals the noise (Figure~\ref{fig:gate}c). The tail
shape therefore carries the effect (details in Appendix~\ref{app:flip}).

\subsection{A recipe for which axis to synthesize along}
\label{sec:recipe}

\begin{algorithm}[!ht]
  \caption{Choosing what to synthesize for a target model}
  \label{alg:recipe}
  \begin{algorithmic}[1]
    \Require target model $M$ with input transform $T$; real sample $X$ from the target modality; generator
      $G$ with parameters $\theta$; threshold $\epsilon_a$ for each axis $a$
    \For{each axis $a \in \{$structure, dimensionality, tail, SNR$\}$}
      \State $s_a \gets$ spread of $a(T(X))$ across real sessions \Comment{no training}
      \State mark $a$ \emph{live} if $s_a > \epsilon_a$, else \emph{dead}
    \EndFor
    \For{each live axis $a$}
      \State find the parameter $\theta_a$ that moves $a(T(G(\theta)))$ across the real range \Comment{in silico}
    \EndFor
    \State pretrain $M$ on $G$ at low, mid, and high $\theta_a$; keep $a$ if transfer tracks it
    \State \Return the best $\theta_a$; then scale the corpus size
  \end{algorithmic}
\end{algorithm}

Our results turn into a procedure for a new target model (Algorithm~\ref{alg:recipe}). Its first two steps
need no training. The axes are measured on real data after the model's own input transform, and a generator
parameter is calibrated in silico. Only the last step trains, with one small sweep. UCR and calcium are two worked
cases. On UCR, the distribution shape survives z-scoring, but the confirming sweep shows that transfer does not
track it, while it does track structure (Sections~\ref{sec:ucr} and~\ref{sec:gate}). On calcium, matching the
shape fixes coupling, so the first step marks structure dead. Varying the event density moves effective rank
only within a narrow band, over which transfer does not track it (Section~\ref{sec:effrank}).

\paragraph{Predictions for the untested modalities.} Applying the principle to the
modalities of Table~\ref{tab:modalities} nominates the likely governing axis without training
(Table~\ref{tab:predictions}). Each prediction assumes the input transform a foundation model for
that modality would typically use. These predictions are falsifiable: each needs only that foundation model
and a matched TransferForge corpus.

\begin{table}[!ht]
  \centering
  \caption{\textbf{Candidate governing axis.} Tail is the excess kurtosis of Table~\ref{tab:modalities}.}
  \label{tab:predictions}
  \begin{tabular}{lccc}
    \toprule
    Modality & Typical input transform & Tail & Candidate governing axis \\
    \midrule
    EEG, fMRI, weather & per-window normalization & $\approx 0$ & structure \\
    spikes             & unnormalized counts      & $3.7$       & distribution shape (weaker) \\
    calcium            & unnormalized $\Delta F/F$ & $12.0$     & distribution shape (shown here) \\
    \bottomrule
  \end{tabular}
\end{table}

\section{Discussion}
\label{sec:discussion}

A pretraining corpus has three properties that can shape transfer: its structure, its dimensionality, and
the distribution shape of each unit's values. Which of them can matter is set by the model's input
pipeline, not by the data. When the model z-scores each input window, the distribution shape is inert and
structure governs, and synthetic data matches a real corpus of equal size and exceeds it once more is
generated. When the model encodes calcium $\Delta F/F$ unnormalized, the distribution shape becomes a governing
constraint, and a shape-matched synthetic corpus replaces real pretraining with no real neural data. Flipping only the input
normalization, with the model and data fixed, moves the distribution-shape manipulation from inert to
active, so the pipeline, not the modality, decides whether it matters.

This changes how pretraining data can be obtained. Instead of collecting more real data and hoping it
transfers, one can measure which properties survive a model's input transform, synthesize a corpus that has
the right ones, and generate as much of it as needed. The first steps need no training, so the choice of
what to synthesize is made before any compute is spent. The gain is largest where real data is scarce,
costly, or private, as neural recordings are. There is no corpus that suits every model, only a corpus
matched to a pipeline.

\clearpage
\subsection*{Reproducibility statement}
This paragraph points to where the information needed to reproduce our results lives. The TransferForge
generator and the calcium distribution-shape transform are specified in Section~\ref{sec:method}, with full
implementation details (GP/SCM construction, GCaMP kernel, event-density $q$, noise floor) in
Appendix~\ref{app:impl}; the axis definitions and their cross-modality validation are in
Appendix~\ref{app:background}. Accuracies for each pretraining corpus, seed counts, and step-/budget-matching are reported with
each experiment in Section~\ref{sec:results}. Code and configuration files will be released as
supplementary material on acceptance.

\subsection*{AI use statement}
The authors used a generative AI coding assistant for implementation scaffolding (data-generation
and figure code), experiment orchestration, and drafting/editing of this manuscript. All code was
reviewed and executed by the authors, all numerical results were produced by the authors' runs, and
the authors take responsibility for the final content.

\bibliographystyle{iclr2027_conference}
\bibliography{refs}

\appendix

\section{Background: the structural axes across modalities}
\label{app:background}

TransferForge did not choose its two structural axes arbitrarily. They are the distillation of an extensive
prior analysis, across the data modalities of Table~\ref{tab:modalities}, of \emph{what} makes data predictable/learnable and
which candidate ``predictability'' metrics are genuine versus confounded. This appendix substantiates the two structural axes used in the main text; it is background that
informs the generator, not a claim of this paper.

\paragraph{The two axes.} We characterize a population $X \in \mathbb{R}^{T \times N}$ ($T$
timepoints, $N$ units) by two quantities. \emph{Embeddedness} (structure axis) is how
reconstructable a unit is from the rest of the population; at the population level we summarize it
by the mean off-diagonal coupling
$\mathrm{coup}(X) = \operatorname{mean}_{i\neq j}|\mathrm{corr}(x_i,x_j)|$, and at the unit level by
the cross-validated $R^2$ of reconstructing a unit from the rest. \emph{Effective rank}
(dimensionality axis) is the participation ratio of the correlation spectrum,
$\mathrm{er}(X) = (\sum_k \lambda_k)^2 / \sum_k \lambda_k^2$: the number of independent modes the
population carries.

\paragraph{The axes measured across modalities.} Computing the axes across modalities
(Table~\ref{tab:modalities}) shows they span a wide plane and that modalities occupy very different
regions. Human macro-scale recordings (task/resting fMRI, EEG) are low-dimensional and
high-coupling, nearly collapsed, so there is little diversity to select from and their intrinsic
predictability is already near ceiling. Cellular-resolution recordings (spikes, calcium) and
broad multivariate sensor data (weather, vision) span far more of the plane: they have a small
embedded scaffold plus a rich private complement, which is exactly the regime where the
\emph{composition} of a pretraining pool can matter. This is why a generator that can place data
anywhere on the plane is useful.

The table also reports the \emph{distribution shape}, summarized by its tail (excess kurtosis) and by the
transient-to-baseline SNR used for calcium distribution shape matching. Both are invariant to the per-series
affine normalization that pipelines like UCR/PatchTST apply, so they are properties of each modality, not
artifacts of preprocessing. Whether the shape \emph{matters} for transfer is set not by its level but by the
model's input pipeline (Section~\ref{sec:marginal}).

\paragraph{Embeddedness is a genuine structural invariant, not an amplitude artifact.} A recurring
trap is that continuous ``predictability'' metrics (kurtosis, forward-prediction $R^2$,
participation ratio) collapse onto signal amplitude/SNR or a recording attribute such as unit
count. Embeddedness survives this scrutiny \emph{as a confound-matched contrast}: (i) on spikes,
with firing rate matched by decile, more-embedded units are more reconstructable than less-embedded
ones (narrow-waveform inhibitory $0.179$ vs.\ wide $0.112$, Wilcoxon $p=1.5\times10^{-20}$);
(ii) a per-unit cross-unit information-theoretic predictability equals coupling
($\rho=+0.763$, $n{=}45{,}593$ units) and \emph{strengthens} to $+0.815$ after partialling out
firing rate and remains $+0.751$ after also partialling out temporal predictability, so it is not
a rate or single-axis rename; (iii) against a ground-truth wiring diagram (\textit{C.\ elegans}
connectome), unit embeddedness aligns with anatomical synaptic degree
($\rho=+0.30$ pooled, sign-consistent in $4/4$ sessions).

As a \emph{free per-unit score} embeddedness stays partly amplitude-entangled (\textit{C.\ elegans}
partial $\rho\!\to\!0.10$), so it is used as an \emph{axis to span}, not a scalar to maximize
(the rank-1 pure-embedded extreme is degenerate; see Section~\ref{sec:ucr}).

\paragraph{The axes trade off, so real data cannot be set freely.} Across $220$ random neuron pools
($30$--$90$ neurons) drawn from the largest \textit{C.\ elegans} whole-brain session \citep{dandi000565}, the two
axes are strongly \emph{negatively} correlated (Pearson $\rho\approx-0.85$ with both axes on the correlation
matrix): more coupling comes with lower effective rank. One therefore cannot obtain, from real recordings alone, a population at
an arbitrary (structure, dimensionality) setting, and most of the plane is empty. This is the gap
TransferForge spans: a generator that places a population off the trade-off diagonal, at a
prescribed structural setting.

\section{Extended related work}
\label{app:related}

\paragraph{Curriculum learning and data selection.} Curriculum learning \citep{bengio2009curriculum}
and data pruning by an example's learning value \citep{sorscher2022beyond,mindermann2022prioritized}
score or reorder \emph{existing} data; we instead identify the properties that make data transfer and
\emph{synthesize} data that has them.

\paragraph{Objective, structure, and dimensionality.} Our objective is masked reconstruction
\citep{devlin2019bert,he2022masked}. Self-supervised representations are known to collapse to a
low-dimensional subspace \citep{chen2021exploring,jing2022collapse}. The $n_{\text{roots}}=1$ extreme of our
dimensionality axis is a low-rank \emph{corpus}, the data-side analogue of that failure; it transfers worse
than the interior optimum (Section~\ref{sec:ucr}), but we do not measure the rank of the learned
representation.

The structure axis rests on population activity concentrating in a few shared,
cell-type-dependent modes
\citep{gao2015simplicity,stringer2019high,churchland2012neural,okun2015diverse,ecker2014state}; the
dimensionality axis on effective rank / participation ratio \citep{roy2007effective,levina2004maximum}
and the link between low intrinsic dimension and better learning
\citep{ansuini2019intrinsic,pope2021intrinsic}, with data quality known to shift scaling
\citep{kaplan2020scaling,hoffmann2022training}. Our neural experiments use population-decoding models
\citep{azabou2023unified,nie2023patchtst} and recast the finding that predictable-first pretraining aids
neural transfer \citep{bae2026decoding} as a statement about where data sits on these axes.

\section{Modality data for Table~\ref{tab:modalities}}
\label{app:modalities}

\begin{table}[!ht]
  \centering
  \caption{\textbf{Data behind Table~\ref{tab:modalities}.} Units and lengths are those used for tail and SNR.
  For spikes and calcium, structure and dimensionality are computed on larger session sets ($24$ IBL
  sessions; $432$ Allen sessions).}
  \label{tab:modality_sources}
  \small
  \begin{tabular}{llllr}
    \toprule
    Modality & Dataset & Unit & Units & Length \\
    \midrule
    fMRI (task)    & HCP EMOTION task \citep{vanessen2013hcp}          & voxel      & $24\times24$ subj.\ & 88 TRs \\
    EEG            & TUH abnormal EEG (TUAB) \citep{obeid2016temple} & channel    & $40\times23$ windows & 2000 samples \\
    fMRI (resting) & HCP S1200 resting, 360 ROIs \citep{vanessen2013hcp} & ROI     & $24\times40$ subj.\ & 1200 TRs \\
    weather        & WEATHER-5K, hourly stations \citep{han2024benchmarking} & station$\times$variable & 2500 & 8760 h \\
    \textit{C.\ elegans} & DANDI:000565, whole-brain calcium \citep{dandi000565} & neuron     & 431 (4 sessions) & 1500 frames \\
    vision         & ImageNet-1k validation, $16\times16$ grayscale \citep{deng2009imagenet} & pixel & 256 & -- \\
    spikes         & IBL Neuropixels, 50\,ms bins \citep{international2025brain} & sorted unit & 23{,}220 & 36{,}000 bins \\
    calcium        & Allen Brain Observatory $\Delta F/F$ \citep{devries2020large} & ROI & 2846 (40 sessions) & 20{,}000 frames \\
    language       & PPMI word--context matrix, Project Gutenberg      & word       & 4000 & -- \\
    \bottomrule
  \end{tabular}
\end{table}

\section{Generator implementation details}
\label{app:impl}
TransferForge draws root nodes from an $O(T)$
AR(1)-smoothed white-noise process with a timescale of $3$ samples. Derived nodes are linear-plus-nonlinear reconstructions
of their parents (structural-causal-model edges over a Gaussian-process root construction), with a
private-noise term $\mathrm{priv}\cdot\sigma(x_j)\,\eta_j$ per derived node that decouples coupling
strength from the embedded fraction. The calcium distribution-shape transform maps a latent to a
calcium-like signal via high-quantile sparse-event thresholding, a double-exponential GCaMP kernel, a
rare-huge-event mixture for the extreme tail, and a baseline-noise floor for SNR; the effective-rank
ablation drives built-window participation ratio through the event-density parameter $q$ while
re-matching excess kurtosis and SNR at each level.

\section{Experimental details}
\label{app:expdetails}

\paragraph{UCR (normalizing pipeline).} We pretrain a PatchTST encoder and probe transfer with a frozen
logistic-regression classifier on the 22-dataset UCR archive, reporting mean accuracy across datasets. The
$n_{\text{roots}}$ sweep uses 3 seeds at a fixed budget (40 epochs, 128 source channels). The
budget-matched comparison against the real forecasting corpus uses $1776$ z-scored length-256 windows, cut
with stride $64$ from economic series long enough for that length, so neighbouring windows overlap. That
real corpus transfers to UCR classification no better than random init.

\paragraph{Neural drifting-gratings decoding (unnormalized pipeline).} We pretrain POYO+ (step-matched across pretraining corpora) and then run a 15-epoch
full finetune (unfrozen encoder) of an 8-way drifting-gratings orientation decoder (chance $0.125$) on a
held-out pool of $139$ real calcium sessions, reporting mean held-out test accuracy. Runs that hit the time
limit were rerun at 12 epochs (listed in the caption of Figure~\ref{fig:calcium_axes}). Each pretraining corpus is run with 3 seeds, except
the three that use scarce real data (5 seeds each, including
chance-stalled runs).

\paragraph{Calcium pretraining corpora.} The scarce-real corpora model a lab with few recordings of its own: the
scarce pool holds $19$ real sessions, a quarter of the $77$ used for full-real pretraining, and the
supplemented corpus adds $50$ calcium-shaped synthetic sessions, step-matched to the others. The calcium- and
Gaussian-synthetic pools share the generator setting and differ in the per-unit transform, which sets the
distribution shape and also lowers coupling (Section~\ref{sec:effrank}). Every model is finetuned on the same
$139$ sessions with the same decoder and trainable capacity, so the spread across pretraining corpora ($0.220$ to $0.407$ at fixed
finetuning) comes from pretraining.

\paragraph{Calcium structural sweeps.} The $n_{\text{roots}}$ sweep spans $\{1,8,16,32,64\}$ under the
fixed matched calcium distribution shape. The dimensionality sweep varies the calcium transform's
event-density $q$, re-matching the distribution shape at each level. Its effective rank is measured at pool
level: for each of the $50$ synthetic sessions, the participation ratio of the singular values of centered
$60$-sample windows, median over $200$ windows, then the median across sessions ($4.8$--$6.2$ across levels).
Real sessions measured the same way span $2.1$--$16.8$ (5th--95th percentile).

\paragraph{Why matching the calcium shape collapses the structure axis.} Two transforms that make the
distribution shape real cause the collapse. The high-quantile sparse-event thresholding needed for the
heavy tail drops coupling $0.89\!\to\!0.18$ (rare co-firing among sparse events), and the baseline-noise
floor needed for the anchor session's SNR ($\approx 6.8$) drops it further to $\approx 0.05$. A zero-noise floor restores
structure but sends SNR to $\sim\!174$, far from real. Matching the distribution shape and spanning the
structure axis are therefore in direct tension.

\section{Additional figures}
\label{app:figures}
The following figures support the calcium results of Sections~\ref{sec:neural} and~\ref{sec:effrank}.

\begin{figure}[!ht]
  \centering
  \includegraphics[width=0.9\linewidth]{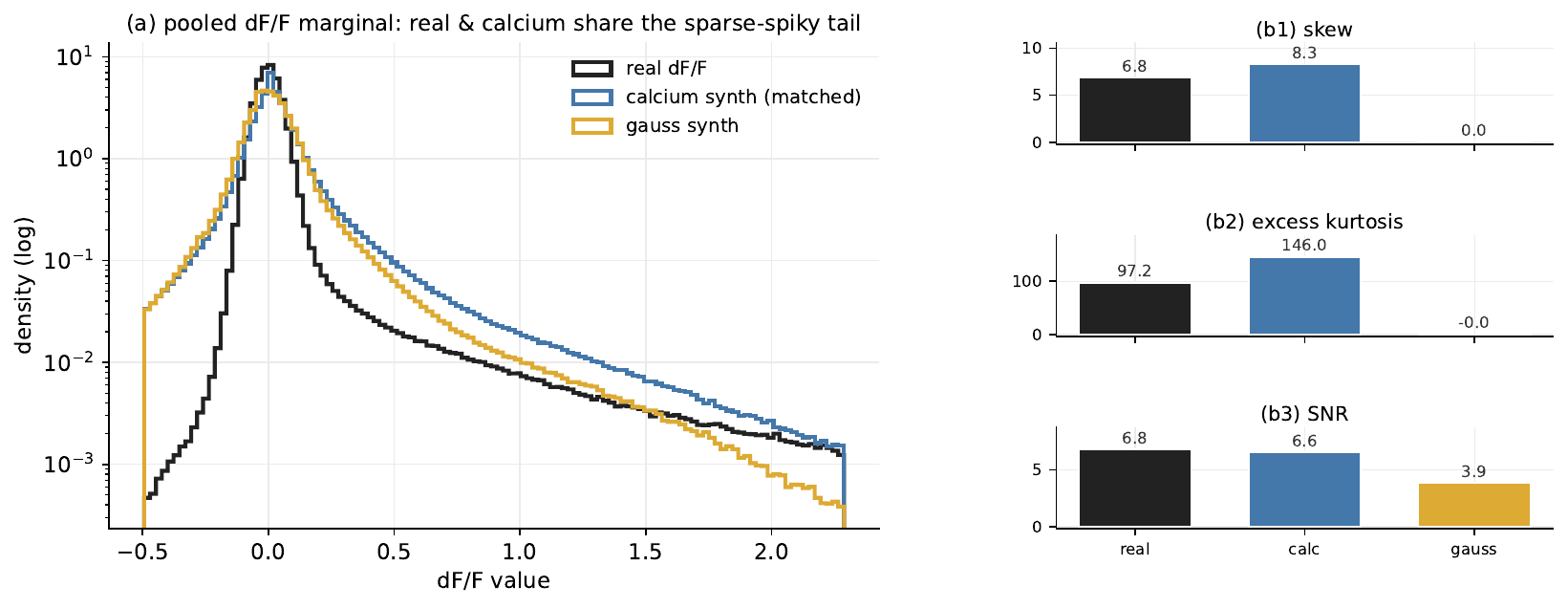}
  \caption{\textbf{Matching the calcium distribution shape.} Left: pooled $\Delta F/F$ value distribution
  (log scale); real and calcium-transformed synthetic share the heavy sparse-spiky tail, while Gaussian
  synthetic does not. Right: per-unit summary statistics (skew, excess kurtosis, SNR); calcium
  synthetic is close to the calibration session on all three, though heavier-tailed, while Gaussian
  synthetic matches none.}
  \label{fig:marginal}
\end{figure}

\begin{figure}[!ht]
  \centering
  \includegraphics[width=0.92\linewidth]{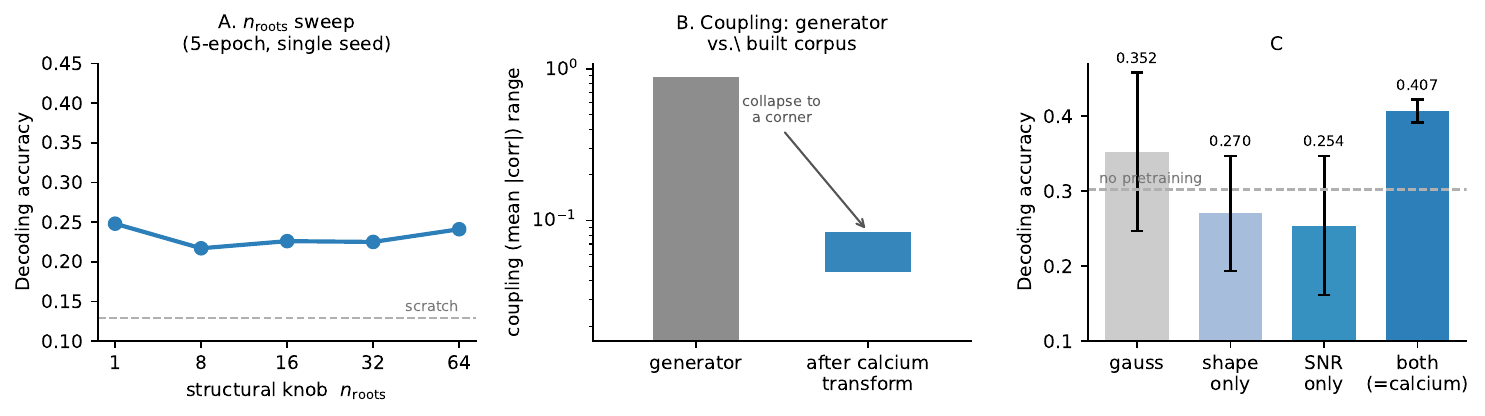}
  \caption{\textbf{Calcium sweeps (drifting-gratings decoding accuracy, chance $0.125$).} (A)~Decoding accuracy vs.\ $n_{\text{roots}}$
  (5-epoch, single seed). (B)~coupling range of the generator vs.\ the built corpus after the calcium
  distribution-shape transform. (C)~Decoding accuracy as the per-unit distribution match is varied
  (gauss / shape-only / SNR-only / both$=$calcium; mean $\pm$ SD over 3 seeds, held-out test accuracy).}
  \label{fig:neural_sweep}
\end{figure}

\section{Normalization flip: further details}
\label{app:flip}

\paragraph{Real-data replication.} Each FRED-MD and exchange-rate series is rank-Gaussianized, preserving
its temporal order while setting its marginal, then warped with the same monotone map across the same
kurtosis range; the identical PatchTST is pretrained both ways (5 seeds, 22 UCR datasets). With
normalization on, the accuracy range is $0.005$, inside seed noise. Without it the spread is $2.9\times$
wider. On this base the sign of the effect is dataset-dependent (a heavier tail mildly helps before turning
down at the extreme), unlike the monotone decline on the synthetic base: normalization decides whether the
manipulation matters, not the direction in which it then moves transfer.

\paragraph{KernelSynth base.} We repeat the same flip on $137$ KernelSynth series \citep{ansari2024chronos}
(length $1024$, cut into $1781$ windows, matching the real base), rank-Gaussianized and warped across the
same kurtosis range (5 seeds, 22 UCR datasets). With normalization on, accuracy stays within $0.811$--$0.816$
(range $0.004$, inside seed noise). Without it the range is $0.009$, $2.0\times$ wider, but close to the
seed spread and not monotone in kurtosis. The invariance under normalization therefore holds for all three
bases (ours, real, KernelSynth); the strength of the activation without it depends on the base.

\paragraph{Why the null under normalization is not a tautology.} Because z-scoring is affine, skewness and
kurtosis reach the model unchanged either way; the null with normalization on is an invariance of the
model, not an erased input. This also explains why UCR never needed distribution-shape matching: its
pipeline makes the shape inert, and its series are near-Gaussian anyway. Calcium is the one modality that is
both strongly non-Gaussian and fed to the model unnormalized. The reverse manipulation, adding per-unit
normalization on the calcium side, is not usable: z-scoring removes the transient amplitude that carries
the drifting-gratings signal, so decoding collapses to chance.

\paragraph{Chronos.} Chronos \citep{ansari2024chronos} scales each series by its mean absolute value and
maps values into uniformly spaced bins, a tokenizer that, unlike per-window z-scoring, keeps each window's
scale relative to its series. We pretrain a Chronos-T5 (\texttt{t5-efficient-tiny}, random init, $3000$ steps) from
scratch on the same warped corpora ($n_{\text{roots}}=4$) and evaluate its native forecasting. The warp is
active: forecasting error rises monotonically with tail heaviness, while a right-skew warp instead helps, a
spread $7\times$ the seed noise (Figure~\ref{fig:gate}b).

\paragraph{Tail shape versus signal-to-noise ratio.} Because Chronos consumes dense series, it separates
the two components of the distribution shape that the calcium window entangles. From the same Gaussian
anchor we sweep tail shape with SNR held clean, and SNR with tail shape held Gaussian. Lowering SNR from
clean down to $2$ leaves forecasting flat; it collapses only at SNR $1$, where the signal variance equals
the injected noise (Figure~\ref{fig:gate}c; medians over 3 seeds). Transfer changes gradually with tail shape, and the
noise floor matters only where it destroys the signal.

\section{Limitations}
\label{app:limitations}

The calcium structural axes are not shown to matter, but structure was never varied at
calcium's input and dimensionality only over pool window effective rank $4.8$--$6.2$, so whether a wider range matters is
open; what separates the default calcium corpus from the effective-rank corpora, its event density or the
training budget, is not isolated; their positive effect is established on UCR (Section~\ref{sec:ucr}). Shape and SNR
separate only on the dense Chronos backbone, not within the calcium window (Appendix~\ref{app:flip}). On the
neural replacement the difference from full-real pretraining is not significant ($5\%$ higher; $p=0.20$,
Welch, 3 seeds); we claim neither superiority nor equivalence. The normalization flip lets both the
distribution shape and each window's scale reach the model, so it does not separate the two; a control that
restores per-window scale with the shape held fixed is left to future work. The neural evidence
rests on a single downstream task, 8-way drifting-gratings orientation decoding; continuous trajectory
decoding and spike-based modalities remain untested.

\end{document}